\RequirePackage{fix-cm}
\documentclass[twocolumn]{svjour3}          % twocolumn
\makeatletter
\if@twocolumn
  \renewcommand\normalsize{%
   \@setfontsize\normalsize\@xpt{12.5pt}%
   \abovedisplayskip=3 mm plus6pt minus 4pt
   \belowdisplayskip=3 mm plus6pt minus 4pt
   \abovedisplayshortskip=0.0 mm plus6pt
   \belowdisplayshortskip=2 mm plus4pt minus 4pt
   \let\@listi\@listI}%

  \renewcommand\small{%
   \@setfontsize\small{8.5pt}\@xpt
   \abovedisplayskip 8.5\p@ \@plus3\p@ \@minus4\p@
   \abovedisplayshortskip \z@ \@plus2\p@
   \belowdisplayshortskip 4\p@ \@plus2\p@ \@minus2\p@
   \def\@listi{\leftmargin\leftmargini
               \parsep 0\p@ \@plus1\p@ \@minus\p@
               \topsep 4\p@ \@plus2\p@ \@minus4\p@
               \itemsep0\p@}%
   \belowdisplayskip \abovedisplayskip}

\else
  \if@smallext
   \renewcommand\normalsize{%
   \@setfontsize\normalsize\@xpt\@xiipt
   \abovedisplayskip=3 mm plus6pt minus 4pt
   \belowdisplayskip=3 mm plus6pt minus 4pt
   \abovedisplayshortskip=0.0 mm plus6pt
   \belowdisplayshortskip=2 mm plus4pt minus 4pt
   \let\@listi\@listI}%

  \renewcommand\small{%
   \@setfontsize\small\@viiipt{9.5pt}%
   \abovedisplayskip 8.5\p@ \@plus3\p@ \@minus4\p@
   \abovedisplayshortskip \z@ \@plus2\p@
   \belowdisplayshortskip 4\p@ \@plus2\p@ \@minus2\p@
   \def\@listi{\leftmargin\leftmargini
               \parsep 0\p@ \@plus1\p@ \@minus\p@
               \topsep 4\p@ \@plus2\p@ \@minus4\p@
               \itemsep0\p@}%
   \belowdisplayskip \abovedisplayskip}
 \else
  \renewcommand\normalsize{%
   \@setfontsize\normalsize{9.5pt}{11.5pt}%
   \abovedisplayskip=3 mm plus6pt minus 4pt
   \belowdisplayskip=3 mm plus6pt minus 4pt
   \abovedisplayshortskip=0.0 mm plus6pt
   \belowdisplayshortskip=2 mm plus4pt minus 4pt
   \let\@listi\@listI}%  

  \renewcommand\small{%
   \@setfontsize\small\@viiipt{9.25pt}%
   \abovedisplayskip 8.5\p@ \@plus3\p@ \@minus4\p@
   \abovedisplayshortskip \z@ \@plus2\p@
   \belowdisplayshortskip 4\p@ \@plus2\p@ \@minus2\p@
   \def\@listi{\leftmargin\leftmargini
               \parsep 0\p@ \@plus1\p@ \@minus\p@
               \topsep 4\p@ \@plus2\p@ \@minus4\p@
               \itemsep0\p@}%
   \belowdisplayskip \abovedisplayskip}
  \fi
\fi

\makeatother
\smartqed  % flush right qed marks, e.g. at end of proof
\usepackage{graphicx}
\usepackage{amsmath}
\usepackage{amssymb}

\usepackage{blindtext}
\usepackage{caption, subcaption}
\usepackage{multirow}
\usepackage{booktabs}
\usepackage{verbatim}
\usepackage{color, colortbl}
\usepackage{xspace, xcolor}
\usepackage{cite}
\usepackage[export]{adjustbox}
\usepackage{comment}
\usepackage{soul}
\usepackage{dsfont}
\definecolor{citecolor}{RGB}{34,139,34}

\usepackage{booktabs}
\usepackage{amsmath}
\usepackage{amssymb}
\usepackage{mathtools}
\usepackage{xspace}
\usepackage{hyperref}
\usepackage{graphicx}
\usepackage{caption}
\usepackage{subcaption}
\usepackage{multirow}
\usepackage{color, colortbl}
\usepackage{fancyvrb}
\usepackage{xcolor}
\usepackage{appendix}
\usepackage{natbib}

\definecolor{Gray}{gray}{0.9}

\usepackage{microtype} 
\usepackage{threeparttable}
\hypersetup{pdfstartview={FitH}}
\hypersetup{pdfborder={0 0 0}}
\hypersetup{pdfpagemode={UseNone}}
\hypersetup{colorlinks=true}
\hypersetup{citecolor=blue}
\hypersetup{linkcolor=blue}

\begin{document}

\title{Focus for Free in Density-Based Counting}
%\title{Point Annotations Beyond Density Maps for Counting}
%\title{Precision Counting: The Art of Repurposing Point Annotation}
%\title{The Power of Repurposed Point Annotations in Counting}
%\title{Repurposing Point Annotation for Robust Counting}

%\title{\st{Counting with Focus for Free}
%\thanks{Grants or other notes
%about the article that should go on the front page should be
%placed here. General acknowledgments should be placed at the end of the article.}
%}
%\subtitle{Do you have a subtitle?\\ If so, write it here}

%\titlerunning{Short form of title}        % if too long for running head

\author{Zenglin~Shi $^{1,2}$       \and
        Pascal~Mettes $^2$       \and
        Cees~G.~M.~Snoek $^2$
}

\authorrunning{Zenglin~Shi et al.} % if too long for running head

\institute{
Zenglin~Shi  \\
\email{shizl@i2r.a-star.edu.sg} \\ \\
Pascal~Mettes  \\
\email{P.S.M.Mettes@uva.nl} \\ \\
Cees~G.~M.~Snoek  \\
\email{cgmsnoek@uva.nl} \\ \\
\small $^1$ I$^2$R, Agency for Science, Technology and Research\\
$^2$ University of Amsterdam
}

\date{Received: date / Accepted: date}
% The correct dates will be entered by the editor

\maketitle
\def\eg{\textit{e.g.}}
\def\ie{\textit{i.e.}}
\def\Eg{\textit{E.g.}}
\def\etal{\textit{et al. }}
\def\etc{\textit{etc.}}
\newcommand{\dimny}{\mathcal{M}\xspace}
\newcommand{\dimyH}{\mathcal{H}\xspace}
\newcommand{\dimyW}{\mathcal{W}\xspace}
\newcommand{\dimH}{\ensuremath{H}}
\newcommand{\dimW}{\ensuremath{W}}
\newcommand{\dimC}{\ensuremath{C}}
\newcommand{\nStage}{\mathcal{L}\xspace}
\newcommand{\scale}{\mathcal{S}\xspace}
\newcommand{\mypartitletwo}[2][2]{\vspace*{-#1 ex}~\\{\noindent {\bf #2}}}
\newcommand{\mypartitle}[1]{\vspace*{-3ex}~\\{\noindent \underline{\bf #1}}}
\newcommand{\todo}[1]{\textcolor{red}{\textbf{#1}}}
\newcommand{\dimn}{\ensuremath{M}}
\newcommand{\apriori}{\textit{a priori}\xspace}
\newcommand{\mapping}{\ensuremath{G}\xspace}
\newcommand{\params}{\ensuremath{\theta}\xspace}
%%% MACROS for the annotations
\newcommand{\data}{\ensuremath{X}\xspace}
\newcommand{\SV}{\ensuremath{X}\xspace}
\newcommand{\pro}{\ensuremath{P}\xspace}
\newcommand{\gt}{\ensuremath{G}\xspace}
\newcommand{\npro}{\ensuremath{N}\xspace}
\newcommand{\featSpace}{\ensuremath{\mathrm{\cal X}}\xspace}
\newcommand{\lSpace}{\ensuremath{\mathrm{\cal Y}}\xspace}
\newcommand{\labbb}{\ensuremath{\mathbf{t}}\xspace}
\newcommand{\state}{\ensuremath{z}\xspace}
\newcommand{\nframes}{\ensuremath{T}\xspace}
\newcommand{\kupdate}{\ensuremath{\boldsymbol{\varphi}}\xspace}
\newcommand{\sol}{\ensuremath{\boldsymbol{\beta}}\xspace}
\newcommand{\nsamples}{\ensuremath{N}\xspace}
\newcommand{\Msamp}{\ensuremath{M_{\mathrm{s}}}\xspace}
\newcommand{\nparticles}{\ensuremath{P}\xspace}

\newcommand{\nDepth}{\ensuremath{D_{\mathrm{max}}}\xspace}
\newcommand{\nTrees}{\ensuremath{K}\xspace}
\newcommand{\Xmat}{\ensuremath{\mathbf{X}}\xspace}
\newcommand{\Ymat}{\ensuremath{\mathbf{Y}}\xspace}
\newcommand{\HH}{\ensuremath{\mathbf{H}}\xspace}
\newcommand{\Smat}{\ensuremath{\mathbf{S}}\xspace}
\newcommand{\Dmat}{\ensuremath{\mathbf{D}}\xspace}
\newcommand{\eye}{\ensuremath{\mathbf{e}}\xspace}
\newcommand{\err}{\ensuremath{\boldsymbol{\xi}}\xspace}
\newcommand{\coeff}{\ensuremath{\mathbf{w}}\xspace}
\newcommand{\samp}{\ensuremath{\mathbf{x}}\xspace}
\newcommand{\laby}{\ensuremath{\mathbf{y}}\xspace}
\newcommand{\func}{\ensuremath{\mathbf{g}}\xspace}
\newcommand{\thresh}{\ensuremath{\tau}\xspace}
\newcommand{\treedepth}{\ensuremath{\Gamma_{\mathrm{depth}}}\xspace}
\newcommand{\sampler}{\emph{Sampler}}
\newcommand{\normal}{\ensuremath{\mathrm{\cal N}}\xspace}
\newcommand{\ssvmCost}{\ensuremath{\ell}\xspace}
\newcommand{\Perp}{\perp \! \! \! \perp}
\def\ci{\perp\!\!\!\perp}
\newcommand{\RR}{I\!\!R}

\newcommand{\labl}{\ensuremath{y}\xspace}

\newcommand{\mean}{\ensuremath{\mu}\xspace}

\newcommand{\Mult}{\ensuremath{\mbox{Mult}}\xspace}
\newcommand{\Cat}{\ensuremath{\mbox{Categorical}}\xspace}
\newcommand{\argmin}{\mathop{\mathrm{arg\,min}}}
\newcommand{\argmax}{\mathop{\mathrm{arg\,max}}}

\newcommand{\cs}[1]{\textcolor{red}{[\textbf{CS}: #1]}}
\newcommand{\psmm}[1]{\textcolor{orange}{[\textbf{PM}: #1]}}
\newcommand{\zl}[1]{\textcolor{blue}{[\textbf{ZL}: #1]}}
\newcommand{\sm}[1]{\textcolor{red}{[\textbf{SM}: #1]}} %Subhransu

\begin{abstract}
This work considers supervised learning to count from images and their corresponding point annotations. Where density-based counting methods typically use the point annotations only to create Gaussian-density maps, which act as the supervision signal, the starting point of this work is that point annotations have counting potential beyond density map generation. We introduce two methods that repurpose the available point annotations to enhance counting performance. The first is a counting-specific augmentation that leverages point annotations to simulate occluded objects in both input and density images to enhance the network's robustness to occlusions. The second method, foreground distillation, generates foreground masks from the point annotations, from which we train an auxiliary network on images with blacked-out backgrounds. By doing so, it learns to extract foreground counting knowledge without interference from the background. These methods can be seamlessly integrated with existing counting advances and are adaptable to different loss functions. We demonstrate complementary effects of the approaches, allowing us to achieve robust counting results even in challenging scenarios such as background clutter, occlusion, and varying crowd densities. Our proposed approach achieves strong counting results on multiple datasets, including ShanghaiTech Part\_A and Part\_B, UCF\_QNRF, JHU-Crowd++, and NWPU-Crowd.
\end{abstract}

\section{Introduction}

The human ability to count is an intriguing cognitive skill that enables us to quantify and comprehend the world. From basic distinctions of ``one'' and ``many'', our counting abilities have evolved into intricate systems like the decimal system. This system equips us with the capacity to represent numbers of infinite magnitude, enabling us to count a plethora of entities, such as species in an ecosystem, stars in the celestial expanse, and buildings in bustling cityscapes. Counting also plays a significant role in computer vision.

In the realm of computer vision, counting objects or identifying the number of specific elements within images or videos is a fundamental task. Computer vision algorithms have employed a variety of approaches to accomplish this feat, such as detection, clustering, and regression. Counting by detection \citep{lin2001estimation,leibe2005pedestrian,wu2007detection,li2008estimating,topkaya2014counting} counts objects by detecting them individually, which is effective when the number of objects is limited, but becomes slow and challenging in crowded scenes. Counting by clustering \citep{brostow2006unsupervised,rabaud2006counting} counts objects by grouping them based on their motion patterns, requiring high frame rates and reliable motion information. Counting by regression \citep{chan2008privacy,chan2009bayesian,chen2012feature,idrees2013multi,chan2011counting} counts objects by learning a direct mapping between image features and a count value, which is suitable for crowded environments and is computationally efficient, but it does not provide the spatial distribution of the objects of interest.

\cite{lempitsky2010learning} propose density maps as a way to incorporate the spatial appearance and constellation of object crowds into counting. Their density-based counting requires just a point annotation per countable object in each training image. These points are smoothed with Gaussian kernels to generate density maps. Counting then becomes a pixel-wise regression problem. Early density-based counting methods \citep{lempitsky2010learning,pham2015count} compute low-level features (\eg, HOG, SIFT) and learn regressors to predict the density maps. Afterwards, deep learning has become dominant in density-based counting, as it predicts density maps in an end-to-end manner. Deep convolutional networks are widely adopted, \eg~ \citep{zhang2016single,onoro2016towards,Li_2018_CVPR,Cao_2018_ECCV,Liu_2018_CVPR_Leveraging, liu2018crowd, liu2019context,jiang2019crowd,sindagi2019ha,xu2022autoscale, xiong2023open} and several recent studies have leveraged the power of vision transformers to further enhance the accuracy of crowd counting models, \eg~\citep{lin2022boosting,sun2021boosting,gao2022congested,yang2022crowdformer,tian2021cctrans,liang2022transcrowd}. 
Advanced loss functions have been proposed to address the limitations of $\ell_1$ and $\ell_2$-norm loss functions in density-based counting, \eg, \citep{Shi_2018_CVPR,shen2018crowd,chan2009bayesian,wang2020distribution,wan2021generalized}. These advanced functions consider outliers, prevent image blurring, and account for local coherence. Some approaches have also leveraged auxiliary information to enhance density regression, \eg~\citep{sam2017switching,kang2020incorporating,sindagi2017generating,ranjan2018iterative,Shi2018vlad,Liu_2018_CVPR_Leveraging,Sam_2018_CVPR,shi2019revisiting,yang2020reverse,liu2021exploiting,zhang2022wide}. While these approaches are effective, they only rely on the point annotations to generate the density maps for training. The premise of this paper is that point annotations serve a purpose for counting beyond creating density maps.

In the conference version of our work \citep{shi2019counting}, we propose two ways to obtain \emph{focus for free}, \ie, free additional supervision signals from the same point annotations. On a local level, we create binary segmentation maps from point annotations and train a segmentation branch to focus only on the regions of interest. We also leverage the relative number of point annotations per image to train a branch with a global density loss to focus on the overall image density.
Since then, several works have used point annotations in different ways to enhance counting. \cite{ma2020learning} propose a scale-aware probabilistic model that leverages the geometric distribution of point annotations to address the scale variation problem. Similarly, \cite{liang2022focal} suggest a focal inverse distance transform map instead of a Gaussian density map to locate objects more accurately in dense regions with overlapping points. \cite{liu2020adaptive} propose learning patch-wise density maps that are created with local count information from point annotations to minimize the discrepancy between training targets and evaluation metrics. In addition, \cite{jiang2020attention, rong2021coarse} use local count information from point annotations to learn attention masks that indicate different density levels of various image regions, reducing the local count error. \cite{modolo2021understanding,qian2022segmentation,jiang2020density} also improve counting by learning an auxiliary segmentation task or density-level classification task by their network architecture design. Instead of a binary segmentation map for attention, \cite{cheng2021decoupled} propose a probability map that shows the likelihood of each pixel being an object, and then learn to predict the probability map as a probabilistic intermediate representation for counting. 

This paper introduces two new ways to repurpose point annotations for free in counting.  
First, in Section~\ref{sec:occlusion} we introduce a counting-specific augmentation that utilizes point annotations to simulate occluded objects in both input and density images to enhance the network's robustness to occlusions. 
Second, in Section~\ref{sec:segmentation} we propose foreground distillation, which generates foreground masks from the point annotations. With these masks we train an auxiliary network on images with blacked-out backgrounds, which learns to distill foreground knowledge without the interference of the background.
In Section~\ref{sec:focus} we show that our new proposals are naturally embedded in our previous local and global approaches \citep{shi2019counting}. Where the counting network in the conference version was based on a convolutional neural network, we integrate all focus-for-free methods in this paper into vision transformers with advanced loss functions, which we detail in Section~\ref{sec:network}.

In Section~\ref{sec:experiments}, our approach is evaluated on five counting datasets. Our results showcase complementary effects of the focus-for-free approaches. When these approaches combined, our models can accurately count objects of interest even in challenging scenarios, such as background clutter, occlusion, and varying crowd densities. Moreover, we achieve strong performance on ShanghaiTech Part\_A and Part\_B, UCF\_QNRF, JHU-Crowd++, and NWPU-Crowd.
Before detailing our approach, we first provide a broader discussion on related work.

%===============================
\section{Related Work}
%===============================

\subsection{Counting with density maps}
The most prevalent technique for object counting in images is to generate density maps through regression. This method was first introduced by \cite{lempitsky2010learning} and has since been the basis of most works. Over time, density-based counting has progressed significantly, thanks to improved network architectures and advanced loss functions. 

\textbf{Network architectures.} 
Deep convolutional networks are widely adopted for counting by estimating density maps from images. Early works, \eg~ \citep{zhang2016single,onoro2016towards,sindagi2017generating}, advocate a multi-column convolutional neural network to encourage different columns to respond to objects at different scales. Despite their success, these types of networks are hard to train due to structure redundancy \citep{Li_2018_CVPR} and conflicts resulting from optimization among different columns \citep{Shen_2018_CVPR, Sam_2018_CVPR}. More recently, single-column deep networks have gained popularity due to their simpler architecture and improved training efficiency, \eg, \citep{Li_2018_CVPR,Cao_2018_ECCV,Liu_2018_CVPR_Leveraging, liu2018crowd, liu2019context,jiang2019crowd,sindagi2019ha,hu2020count, wang2022eccnas}. For example, \cite{Li_2018_CVPR} combine a VGG network with dilated convolution layers to capture multi-scale contextual information. \cite{sindagi2019ha} propose a hierarchical attention-based network with attention mechanisms at various levels to enhance network features selectively. \cite{hu2020count} propose an encoder-decoder network with neural architecture search techniques. 

Several recent studies have leveraged the power of vision transformers to enhance the accuracy of crowd counting models, \eg, \citep{lin2022boosting,sun2021boosting,gao2022congested,yang2022crowdformer,tian2021cctrans,liang2022transcrowd}. \cite{lin2022boosting} have incorporated global attention, learnable local attention, and instance attention into their counting model, by combining vanilla vision transformers and convolutional networks. \cite{sun2021boosting} introduces a global context learnable token to guide the counting. \cite{gao2022congested} enhanced large-range contextual information by using a dilated Swin Transformer backbone and feature pyramid networks decoder. To model human top-down visual perception mechanisms, \cite{yang2022crowdformer} have proposed an overlap patching transformer block. \cite{tian2021cctrans} adopt a pyramid transformer and a multi-scale regression head to achieve improved counting performance.

\textbf{Loss functions.}
$\ell_1$- and $\ell_2$-norm loss functions are widely used to measure the per-pixel differences between estimated and ground-truth density maps for network training, but they suffer from problems such as sensitivity to outliers and image blur, pixel independent assumption neglecting the local coherence, and spatial correlation in density maps. Therefore, \cite{Shi_2018_CVPR} propose a negative correlation loss to increase the robustness against outliers. \cite{Cao_2018_ECCV} add the local pattern consistency loss to reduce the sensitivity of $\ell_1$- and $\ell_2$-norm loss. \cite{shen2018crowd} introduce an adversarial loss to make the blurring density maps sharp. \cite{chan2009bayesian} propose a Bayesian loss by calculating class conditional distributions for each annotated points rather than generating discrete density map as supervision. However, Bayesian loss cannot well handle false positives in the background, and requires a special design for the background region. \cite{wang2020distribution} propose an advanced loss function by computing the optimal transport distance between predicted density maps and the ground-truth point maps. \cite{wan2021generalized} further propose an unbalanced optimal transport loss function to preserve the count of the predicted densities and annotated points. Different from pixel-wise $\ell_1$- and $\ell_2$-norm loss, optimal transport loss produces penalties by considering all nearby pixels according to the distances rather than the pixel itself. Therefore, such losses better exploit the position information of the point annotations to provide high-quality supervision. However, optimal transport loss and its variants utilize the Sinkhorn algorithm \citep{peyre2019computational} to obtain the optimal transport matrix, which requires a number of iterations and are carried out in each training step, leading to inefficient training.

This work strives to improve counting by repurposing point annotations, which is compatible with any network architecture and loss function. We show that our approach can work well with both convolutional networks and vision transformers, supervised by an $\ell_1$-norm loss or a distribution matching loss. 

\subsection{Learning auxiliary tasks for counting}
Several approaches have been proposed to enhance density regression by leveraging auxiliary information, \eg~\citep{sam2017switching,sindagi2017generating,ranjan2018iterative,Shi2018vlad,Liu_2018_CVPR_Leveraging,Shen_2018_CVPR,Sam_2018_CVPR,Liu_2018_CVPR_decidenet,yan2019perspective,zhao2019leveraging,shi2019revisiting,yang2020reverse,liu2021exploiting}. \cite{sam2017switching}, for example, train a classifier to select the optimal regressor from multiple independent regressors for particular input patches. \cite{ranjan2018iterative} utilize one network to predict a high-resolution density map and another network to predict a low-resolution density map to help improve the accuracy of crowd counting. \cite{zhao2019leveraging} propose a method involving three heterogeneous attributes, including geometric, semantic, and numeric attributes, as auxiliary tasks to assist in the crowd counting task. \cite{zhao2019leveraging} learn an auxiliary perspective map prediction task to address the issue of scale variation in counting. \cite{liu2021exploiting} developed a novel multi-experts training framework for crowd counting that exploits relations within samples.

In this paper, we also investigate counting from a multi-task perspective, but from a different point of view. We posit that the point annotations serve more purposes than just constructing density maps, and we introduce occlusion simulation, foreground distillation, and focus-for-free \citep{shi2019counting}. Occlusion augmentation leverages point annotations to simulate occluded objects in both the input and density images, enabling the network to become more robust to occlusions. Foreground distillation generates foreground masks from point annotations, and uses the masks to distill foreground counting knowledge, which effectively reduces the impact of background pixels on counting accuracy. Focus-for-free outlines supervised focus from segmentation and global density classification to repurpose the point annotations for free.

\section{Focus for Free: Repurpose Points to Count}
\label{sec:method}
\begin{figure*}[t!]
\centering
\includegraphics[width=0.99\linewidth]{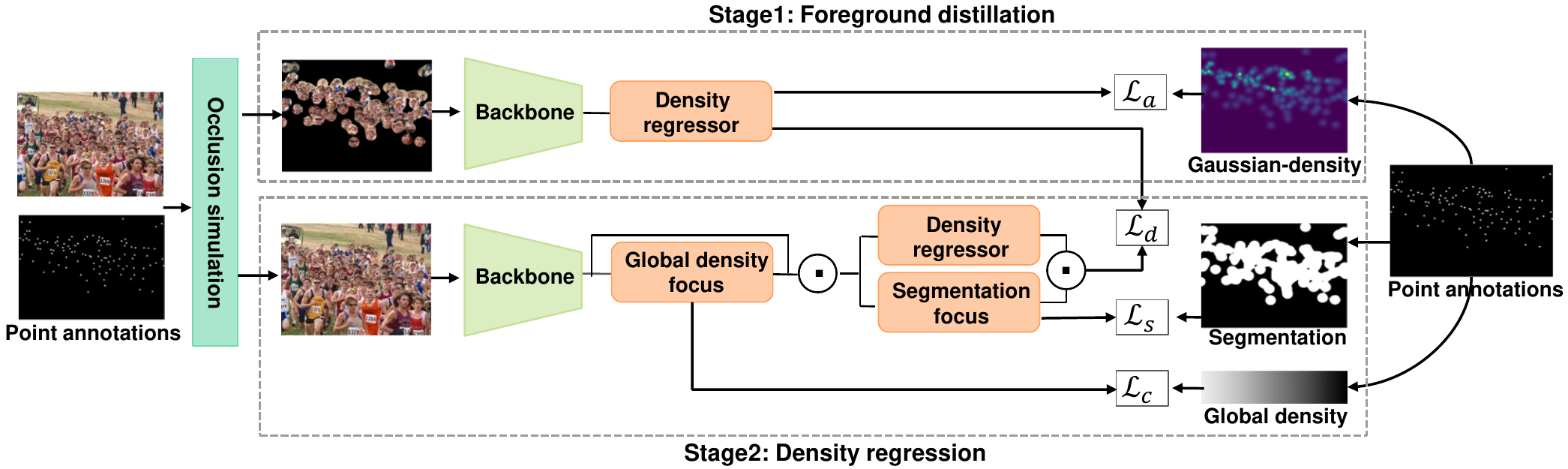}
\caption{\textbf{Overview of our approach.}
We first repurpose the point annotations to simulate occlusion in images and density maps, improving robustness to crowded scenes (Section \ref{sec:occlusion}). Then we adopt a two-stage learning process, incorporating occlusion augmentation. In the first stage, indicated by the top dashed box, we generate foreground masks from the point annotations from which we train an auxiliary network on images with blacked-out backgrounds (Section \ref{sec:segmentation}). We employ a loss function $\mathcal{L}_a$ to supervise the training process, in which the Gaussian-density map is generated from the point annotations. In the second stage, depicted by the bottom dashed box, we transfer the knowledge acquired by the auxiliary network to the density estimation network via distillation. This transfer is guided by a loss function $\mathcal{L}_d$. To ensure that the density estimation network focuses on critical features, we utilize the repurposed point annotations to create a segmentation focus supervised with a segmentation loss $\mathcal{L}_s$ and a global density focus supervised with a global density classification loss (Section \ref{sec:focus}). Finally, we seamlessly integrate all these approaches into the vision transformer backbone, allowing for effective cooperation between the different components. Furthermore, our framework is flexible in accommodating different loss functions (Section \ref{sec:network}).
}
\label{fig:focus-overall}    
\end{figure*}

We formulate the counting task as a density map regression problem. Density-based counting relies on point annotations indicating the locations of objects in training images. These annotations are a set of coordinates $\mathcal{P}$, where each point $P{=}(x,y) \in \mathcal{P}$ corresponds to an object's location. From the point annotations a Gaussian-density map is created through convolutions with a normalized Gaussian kernel:
\begin{equation}
D(p)=\sum_{i=1}^N\mathcal{N}\big(p|\mu=P,\sigma_P^2\big),
\label{eq:density}
\end{equation}
where $p$ denotes a pixel location, and $\mathcal{N}(\cdot)$ is a normalized Gaussian kernel with mean $P$ and an isotropic variance $\sigma_P$. Determining $\sigma_P$ for each point $P$ is difficult because of object-size variations caused by perspective distortions. A solution is to estimate the radius $r$ of an object as a function of the $K$ nearest neighbor annotations, as done in the Geometry-Adaptive estimation of \cite{zhang2016single}. 

Given a density map, we optimize a neural network $f$ to learn how to map input images $I_i$ to their output density maps $D_i$, forming a dataset ${(I_i,D_i)}_{i=1}^{N}$. During inference, we obtain the global object count $T_i$ of an input image $I_i$ by summing all the pixel values within the predicted density map $D'_i {=} f(I_i)$, as follows: $T_i {=} \sum{p \in I_i} D'_i(p)$. This operation integrates the predicted density values across the image domain and provides an estimate of the number of objects present in the image.

We will now present three methods that enhance counting by repurposing existing point annotations beyond creating Gaussian-density maps. We begin by introducing a counting-specific augmentation that leverages point annotations to simulate occluded objects in both input and density images. This augmentation improves the network's ability to handle occlusions, as discussed in Section~\ref{sec:occlusion}. Then, we propose foreground distillation in Section~\ref{sec:segmentation}. This method involves generating foreground masks based on point annotations and utilizing these masks to distill foreground knowledge. By incorporating this approach, we reduce the impact of background pixels on counting accuracy. To ensure that the network focuses on critical features during density regression, we introduce both local and global focus with explicit supervision derived from point annotations in Section~\ref{sec:focus}. Finally, we integrate all the aforementioned methods into vision transformers with advanced loss functions, which are detailed in Section~\ref{sec:network}. A comprehensive overview of our method is provided in Figure~\ref{fig:focus-overall}.

\subsection{Counting-specific occlusion simulation}
\label{sec:occlusion}
In counting, we often deal with congested scenes and many objects, which leads to occlusions.
To address this challenge, we reuse annotation points to augment overlapping objects. Occlusions often occur when objects overlap, making it difficult to count them accurately. One approach to handle different levels of occlusion is to train a network with a sufficient number of samples for each level. However, the distribution of occlusion levels in existing benchmarks is long-tailed, meaning there are not enough images to learn from natural occlusions \citep{zhang2016single,idrees2018composition,sindagi2020jhu-crowd++,guerrero2015extremely}. To address this issue, we suggest simulating various occlusion levels by using annotation points. This will enable us to train the network on a more diverse set of occlusion levels, improving its ability to identify and count overlapping objects, even in congested scenes.

\textbf{Occlusion simulation.} 
Given a training sample consisting of an input image and its point annotations, $(I, \{P_i\}_{i=1}^{N})$, we represent any object $\mathcal{O}_i$ in the image $I$ as $(x_i,y_i,2\sigma_i)$. Here, $(x_i, y_i)$ represents the coordinate of the corresponding annotated point $P_i$, and $\sigma_i$ is the variance of the corresponding Gaussian kernel. This representation assumes that the object $\mathcal{O}_i$ is circular and its radius is approximately equal to $2\sigma_i$, as proposed in~\cite{lempitsky2010learning}. To generate occlusion scenarios, we randomly select an object, $\mathcal{O}_{occ}$, to be occluded. We then select one of its neighboring objects, denoted as $\mathcal{O}_{copy}$, which will be copied and pasted to occlude $\mathcal{O}_{occ}$. The pasted position of $\mathcal{O}_{copy}$ determines how $\mathcal{O}_{occ}$ will be occluded, defined as: 
\begin{equation}
  \begin{aligned}
  x_{paste} &= \lfloor r \cdot cos(\theta) \rfloor + x_{occ},\\
  y_{paste} &= \lfloor r \cdot sin(\theta) \rfloor + y_{occ},
  \end{aligned}
  \label{eq:occlusion}
\end{equation}
where $r{=}r_{copy}+r_{occ} \epsilon_r$ and $\theta {=} 2\pi \epsilon_{\theta}$. $\epsilon_r$ and $\epsilon_{\theta}$ are randomly sampled from $\mathcal{U}(0,1)$. $\lfloor \cdot \rfloor$ is the floor operator. $r$ and $\theta$ decide how much and where $\mathcal{O}_{occ}$ will be occluded. 

\textbf{Blending.} Directly pasting objects on an image creates boundary artifacts, which may affect the network's learning ability. To handle boundary artifacts, we perform blending to smooth out the boundary artifacts. Specifically, we first copy and paste the object $\mathcal{O}_{copy}$ to the position $(x_{paste},y_{paste})$ of the image $I$, generating a new image $I'$. Then, we compute the binary mask $\alpha$ of the pasted object $\mathcal{O}_{paste}$, which we represent by $(x_{paste},y_{paste},r_{paste})$ where $r_{paste}{=}r_{copy}$. To smooth out the edges of the pasted object, we apply a Gaussian filter to the binary mask $\alpha$ and obtain a smoothed mask $\tilde \alpha$. From $\tilde \alpha$ we construct the occluded image $I_{occ}$ and its corresponding ground-truth density map:
\begin{equation}
  \begin{aligned}
  I_{occ} &= (1-\tilde \alpha)\odot I + \tilde \alpha \odot I',\\
  D_{occ} &= D + G(\mathcal{O}_{paste}),
  \end{aligned}
  \label{eq:blending}
\end{equation}
where $D$ denotes the ground-truth density map of the image $I$, $G(\mathcal{O}_{paste})$ denotes the density map of the pasted object $\mathcal{O}_{paste}$, and $G$ denotes the Gaussian kernel. Equation \ref{eq:blending} states that the pixels to be occluded in the original image are replaced by the pixels of the pasted object to obtain the occluded image. The corresponding occluded density map is obtained by simply adding the density map of the pasted object to the original density map. 

\textbf{Occluding adaptively.} For a training image, we need to decide how many of its objects will be occluded with our approach. A simple way is to set a fixed ratio. However, the training image may already have naturally occluded objects. Thus, we should create new occlusion for the training image according to its current occlusion level. Intuitively, the lower the occlusion level, the higher the amount of occlusions that should be simulated. To compute the occlusion level, we create an occlusion map $M$ for the training image by,
\begin{equation}
\begin{aligned}
&M(x,y) = \sum_{i=1}^N S(x,y;\mathcal{O}_i), \text{where} \\ &S(x,y;\mathcal{O}_i) = \mathds{1}(||x - x_i||^2 + ||y - y_i||^2 \leq (r_{i})^2). 
\end{aligned} 
\label{eq:occmap} 
\end{equation}
Here $(x,y)$ denotes a pixel location, $\mathcal{O}_i{=}(x_i,y_i,r_{i})$ denotes an object, and $\mathds{1}(\cdot)$ is a binary indicator function, which states that a pixel obtains a value of one if it is within an object region. Hence, each pixel value in the occlusion map $M$ indicates the amount of object regions it belongs to. The occlusion level $\bar M$ is computed as the average value of non-zero pixels in the occlusion map $M$. Then the percentage of the objects to be occluded can be adaptively obtained by $\beta $/$\bar M$ where $\beta$ is the upper boundary, and $\beta{=}0.3$ empirically performs well.
\begin{figure}[t!]
\centering
\begin{subfigure}{\linewidth}
\includegraphics[width=\textwidth]{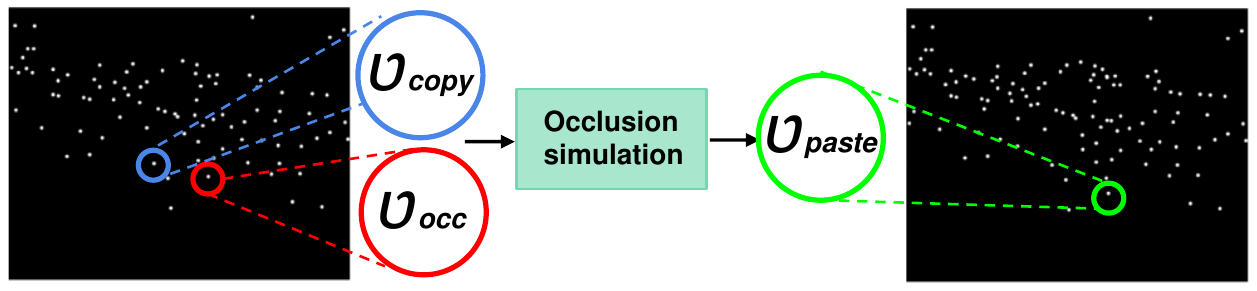}
\caption{\textbf{Occlusion simulation for point annotations}}
\label{fig:augmentation-a}    
\end{subfigure}
\begin{subfigure}{\linewidth}
\includegraphics[width=\textwidth]{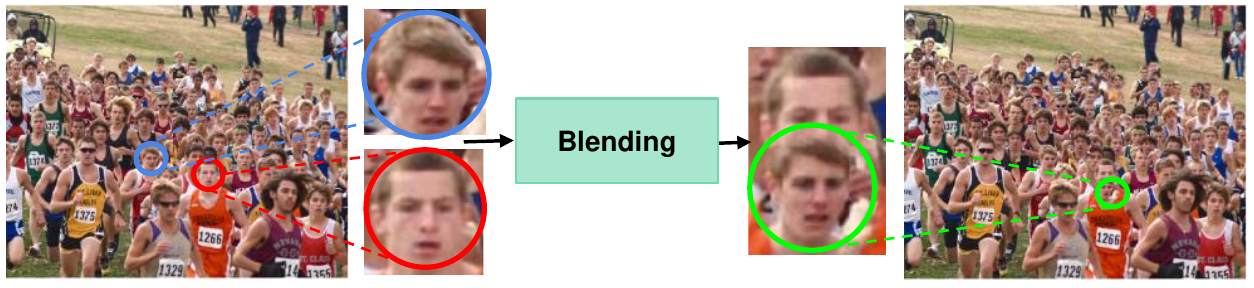}
\caption{\textbf{Occlusion simulation for image}}
\label{fig:augmentation-b}    
\end{subfigure}
\caption{\textbf{Occlusion simulation.} (a): We randomly choose two object points, one for the object which will be occluded, denoted by $\mathcal{O}_{occ}$ and another for the object which will be copied and pasted, denoted by $\mathcal{O}_{copy}$. The pasted position of $\mathcal{O}_{copy}$ is computed based on Eq. (\ref{eq:occlusion}), as indicated by $\mathcal{O}_{paste}$.
(b): We generate an occluded image with $\mathcal{O}_{occ}$, $\mathcal{O}_{copy}$, and $\mathcal{O}_{paste}$. To avoid boundary artifacts, we perform blending to smooth out the boundary artifacts.
}
\label{fig:cases}    
\end{figure}

\subsection{Foreground distillation}
\label{sec:segmentation}
For the second way to repurpose point annotations, we introduce foreground distillation to obtain better density maps without the impact of background.
Typically, a network for density map prediction is learned by optimizing a loss function of the form
$\mathcal{L}_d{=} \ell_p(f_d(I)-D)$
where $f_d$ the network, $I$ the input image, $D$ the Gaussian-density map, and $\ell_p$ the Lp norm loss function. We reformulate the loss function by dividing the density map into two components, one representing the background of the input image and the other corresponding to the foreground containing the objects of interest:
$\mathcal{L}_d{=} \ell_p({f_d(I)}_{bg}) + \ell_p({f_d(I)}_{fg}-D_{fg})$, where $D_{bg}$ is naturally omitted as every single element in it is zero. Based on this reformulation, we identify two issues with current approaches. First, when $\ell_p({f_d(I)}_{bg})$ is not optimized perfectly, they are prone to misidentifying the background as objects (densities). Second, while trying to transform objects into densities, they cannot fully utilize their learning capacity as they predict densities by ${f_d(I)}_{fg}$ not only on objects but also on the background.

A common approach to mitigate the impact of background pixels is to train a separate segmentation network that differentiate between foreground and background regions. However, achieving a perfect foreground segmentation can be difficult, which may compromise the effectiveness of this approach. To remedy this issue, we propose to further distill foreground counting knowledge. Specifically, we introduce an auxiliary network that is trained on images where the background is blacked-out. This enables the network to learn the intrinsic features of the foreground and reduces the influence of background pixels. The training loss is given as:
\begin{equation}
\label{eq:auxiliary}
  \mathcal{L}_a{=} \ell_p \big({f_a(I\odot S)}-D\big),
\end{equation}
where $f_a$ is the auxiliary network, $D$ is the Gaussian-density map, $I\odot S$ is the image with background areas blacked-out, and $S$ is the foreground mask, which is derived as a function of the point annotations and their estimated variance. The binary value for each pixel location $p$ is determined as:
\begin{equation}
S(p) =
\begin{cases}
1 & \text{if  $\exists_{P \in \mathcal{P}} \big( ||p - P||^2 \leq \sigma_P^2$} \big),\\
0 & \text{otherwise.}
\end{cases}
\label{eq:segfocus}
\end{equation}
Equation~\ref{eq:segfocus} states that a pixel $p$ obtains a value of one if at least one point $P$ is within its variance range $\sigma_P$ as specified by a kernel estimator.

The optimization of the auxiliary network will be stopped when it achieves maximum accuracy on the validation set. Then we distill the knowledge of the trained auxiliary network to the density prediction network with the loss:
\begin{equation}
\label{eq:distillation}
  \mathcal{L}_d{=} \ell_p \big(D' - f_a(I\odot S)\big),
\end{equation}
where $D'$ represents the predicted density map. Unlike traditional distillation methods that compress information \citep{hinton2015distilling}, our distillation technique aims to extract foreground counting knowledge. In the density prediction loss function $\mathcal{L}_d$, we replace the Gaussian density map with our distilled density map for the final training, rather than do both. Our approach can be considered independently or in combination with traditional segmentation-based methods to reduce the impact of the background.

\begin{figure*}[t!]
\centering
\includegraphics[width=0.8\linewidth]{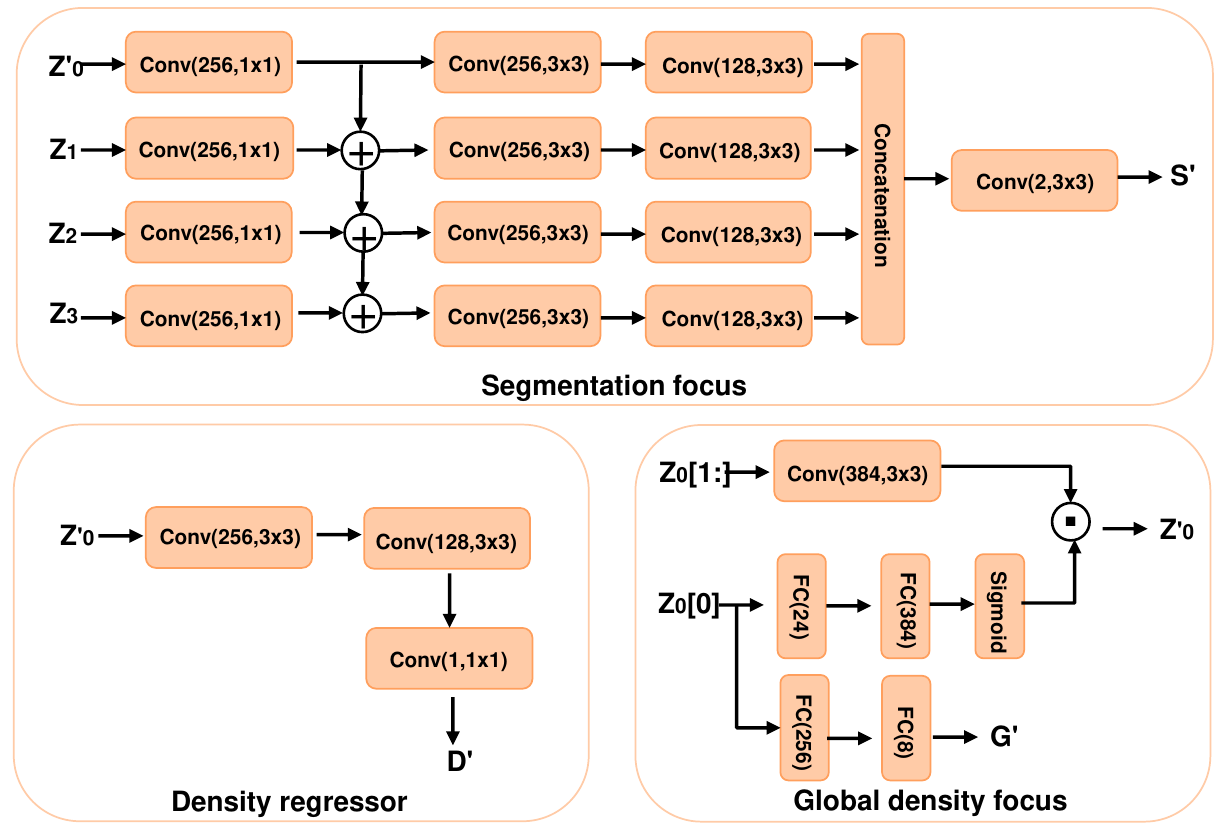}
\caption{\textbf{The network architectures} of density regressor, global density focus, and segmentation focus. The convolution layer is denoted as Conv(N, K×K), where N represents the number of channels and K×K represents the kernel size.  The fully connected layer is represented as FC(M), where M indicates the number of nodes. Following each convolution layer and fully connected layer, there is a ReLU activation layer applied. Element-wise addition and multiplication are denoted by the symbols $\bigoplus$ and $\bigodot$, respectively. ${Z_0,Z_1,Z_2,Z_3}$ are the outputs from different layers in the transformer encoder. D', S', and G' denotes the predicted density map, segmentation map, and global density, respectively. 
}
\label{fig:focus-modules}   
\end{figure*}
\subsection{Local and global focus}
\label{sec:focus}
Here, we outline the original global and local focus for free, which are complementary to the occlusion and distillation approaches.

\textbf{Local segmentation focus.} Intuitively, pixels within a specific range of any point annotation should be of high focus, while pixels in undesired regions should be mostly disregarded. However, in the standard optimization setup where the density map is the sole factor determining the loss, each pixel is treated equally. As a result, irrelevant pixels dominate the loss since only a fraction are near-point annotations. To overcome this limitation, we reuse the point annotations to create a binary segmentation map and exploit this map to provide focused supervision through a standalone loss per-pixel weighted focal loss:
\begin{equation}
\mathcal{L}_s = 
 \sum_{{l} \in \{0,1\}} -\alpha{^l} S^l (1-S')^{\gamma_s} log(S'),
\label{eq:focus-seg}
\end{equation}
where $S$ is the segmentation ground-truth created according to Eq. \ref{eq:segfocus}. $\alpha{^l}=1-\frac{|S^l|}{|S|}$ and $S'{=}f_s(I)$ represents the segmentation map predicted by a branch network $f_s$. The focal parameter $\gamma_s$ is set to 2 throughout this network, as recommended by \cite{lin2017focal}.

We utilize the predicted segmentation map $f_s(I)$ to highlight the desired foreground areas while suppressing false background areas in the output density map. This is done by incorporating the mask into the density prediction loss function. Specifically, through the following masked loss function
\begin{equation}
\mathcal{L}_d {=} \ell_p \big(f_d(I)\odot{f_s(I)}-D\big),
\end{equation}
where $\odot$ denotes element-wise multiplication. This approach reduces background errors and improves density prediction.

\textbf{Global density focus.} Aside from providing a local focus through foreground segmentation, point annotations can also be repurposed to achieve a global focus by examining contextual information through global density classification. In dense prediction tasks, it is crucial to comprehend and leverage contextual information. In the context of per-pixel density prediction for counting, global context is associated with the overall crowd density in the image. To capture such global context, we learn a global density classification task. We then utilize this captured global context to develop a global focus that highlights feature maps that align with the global density reflected by the annotation points, leading to improved counting performance.

To perform global density classification, we first obtain a global density ground-truth from the point annotation. Specifically, the global density for a patch $j$ in the training image $i$ is given as:
\begin{equation}
G_{j,i} = \frac{|\mathcal{P}_{j,i}|}{L}, L = \left \lfloor{ \max_{i=1,..,N} \Big (\frac{|\mathcal{P}_{i}|}{Z_i} \cdot Z_{j,i}\Big) / M}\right \rfloor + 1.
\end{equation}
Here $|\mathcal{P}_{j,i}|$ denotes the number of object points in patch $j$, $L$ denotes the global density step size, $Z_i$ and $Z_{j,i}$ denote the number of pixels in image $i$ and patch $j$ respectively. The step size calculation involves determining the highest global density across image patches, while the parameter $M$ denotes the number of global density levels used in the process. By leveraging the global density ground-truth, we train a network $f_c$ for classifying global density with the following loss:
\begin{equation}
\mathcal{L}_c = 
\sum_{{l} \in \{0,1,..,M\}} -G^l (1-G')^{\gamma_c} log(G'),
\label{eq:global density}
\end{equation}
where $G'{=}f_c(I)$ and $\gamma_c$ is set to 2. The $f_c$ model comprises a backbone network and a classifier. The backbone network is shared with the per-pixel density prediction network $f_d$. After capturing the global context, we develop a global focus that emphasizes feature maps aligned with the global density. To achieve this, we employ a new branch network $f_g$ to generate a global density focus output. The resulting focus output is then utilized by the density prediction network's backbone to highlight feature maps that exhibit similar global contextual patterns to the ground-truth density maps. 

\subsection{Network and optimization}
\label{sec:network}
\textbf{Counting transformer.} Vision transformer architectures have shown significant promise in counting and are quickly becoming the preferred backbone. This has motivated us to integrate our methods into transformers, capitalizing on their capabilities to advance counting models.

For the transformer encoder, we adopt the progressive tokenization module introduced by \cite{yuan2021tokens} instead of relying on the basic tokenization approach employed in Vit \citep{dosovitskiy2020image}. This tokenization module enables us to aggregate adjacent tokens into a single token, thereby incorporating local structural information from surrounding tokens and reducing the token length in a step-by-step manner. To be more specific, we utilize a Token-to-Token (T2T) mechanism, where tokens produced by a transformer layer are first reconstructed as an image, then split into overlapping patches, and finally aggregated by flattening the patches. This process allows the local structure of surrounding patches to be embedded into the tokens, which are subsequently fed into the next transformer layer. By performing T2T iteratively, the local structure is integrated into the tokens, and the token length is reduced via the aggregation process. We use an efficient deep-narrow backbone structure to learn the representations from these tokens. A decoder, consisting of three convolution layers, is employed as a density regressor to estimate the density map. 

The decoder for segmentation focus is designed by adopting a multi-level feature aggregation approach. We begin by selecting four layers from the encoder, denoted by $\{z_0, z_1, z_2, z_3\}$ where $z_0$ represents the last layer of the encoder. To emphasize specific selected layers, we deploy four independent streams, each dedicated to processing one of the chosen layers. Within each stream, we utilize a 3-layer convolutional network to process the corresponding features. To encourage interactions between different streams, we introduce a top-down aggregation design through element-wise addition after the first layer. Subsequently, after the third layer, we obtain the fused feature by concatenating the outputs of all streams along the channel dimension. This fused feature is then fed into a convolution layer to produce the segmentation map.

In the case of the global density focus, our decoder comprises three branches. We divide the patch tokens obtained from $z_0$ into a global density class token $z_0[0]$ and the remaining patch tokens $z_0[1:]$. The first branch processes the patch tokens $z_0[1:]$ using a convolution layer for re-weighting. The second branch generates a global density focus map using two fully-connected layers followed by a Sigmoid activation layer, with the global density class token $z_0[0]$ serving as input. Meanwhile, the third branch leverages the global density class token $z_0[0]$ to predict the global density by employing two fully-connected layers, allowing $z_0[0]$ to capture global context information. The generated focus map by the second branch is then utilized to refine the local feature map from the first branch. This refinement process enables the network to emphasize critical feature channels, leading to enhanced features. Fig. \ref{fig:focus-modules} provides the details of the three decoders.

\textbf{Composite loss.} 
The training of the final counting network is carried out in two distinct stages. Initially, an auxiliary network is trained to perform foreground distillation by minimizing the loss function $\mathcal{L}_a$ (as defined in Equation \ref{eq:auxiliary}). In the second stage, we incorporate focus for free into the counting network by training it using a composite loss function. This loss function comprises three distinct components: the distillation loss $\mathcal{L}_d$ (defined in Equation \ref{eq:distillation}), the segmentation loss $\mathcal{L}_s$ (defined in Equation \ref{eq:focus-seg}), and the global density classification loss $\mathcal{L}_c$ (defined in Equation \ref{eq:global density}), \ie, 
\begin{equation}
\mathcal{L} = \mathcal{L}_d + \lambda_s \mathcal{L}_s + \lambda_c \mathcal{L}_c,
\label{eq:final-loss}
\end{equation}
where $(\lambda_s,\lambda_c)$ denotes the weighting parameters of the different loss functions. Occlusion augmentation is used in both stages. We can also use a distribution matching loss \citep{wang2020distribution} for $\mathcal{L}_d$ to obtain an improved result, where $\mathcal{L}_d$ is defined as:
\begin{equation}
\mathcal{L}_d =  \ell_C(D',D)+ \lambda_{ot} \ell_{OT}(D',D)+ \lambda_{tv} \ell_{TV}(D',\hat{D}).
\label{eq:final-loss}
\end{equation}
Here, $\ell_C$ is used to minimize the difference of the total count between the predicted density map $D'$ and the annotated point density map $D$. $\ell_{OT}$ is used to minimize the divergence between $D'$ and $D$ by regarding the density map as a probability map. $\ell_{TV}$ is used to help $\ell_{OT}$ to handle the low-density areas of the crowd. Since the low-density areas are often impacted by the background, we use our distilled density map $\hat{D}{=}f_a(I\odot S)$ instead of $D$ to distill foreground counting knowledge. $(\lambda_{ot},\lambda_{tv})$ are the weights for the corresponding loss term. Throughout this work, these parameters are set to $(0.1,0.01)$, following \cite{wang2020distribution}.

\section{Experiments and results}
\label{sec:experiments}
We first describe the experimental setup. We then analyze the effect of repurposing point annotations proposed in this work, comparing each to their respective baseline. Finally, we report the results of our final approach and compare them to the state-of-the-art.

\begin{table*}[!t]
\renewcommand{\arraystretch}{1.2}
\centering
\resizebox{0.99\linewidth}{!}{
\begin{threeparttable}
\begin{tabular}{@{}lccccccccccccccc@{}}
\toprule
& \multicolumn{6}{c}{\textbf{\textbf{ T2T-Vit}}} & \multicolumn{6}{c}{\textbf{\textbf{CSRNet+}}}\\
\cmidrule(lr){2-7} \cmidrule(lr){8-13}
& \multicolumn{3}{c}{\textbf{Part\_A}} &
\multicolumn{3}{c}{\textbf{UCF\_QNRF}} & \multicolumn{3}{c}{\textbf{Part\_A}} &
\multicolumn{3}{c}{\textbf{UCF\_QNRF}} \\
\cmidrule(lr){2-4} \cmidrule(lr){5-7} \cmidrule(lr){8-10} \cmidrule(lr){11-13} 
& \textit{Low}  & \textit{High} &\textit{Overall}  & \textit{Low}  & \textit{High} & \textit{Overall}& \textit{Low}  & \textit{High} &\textit{Overall}  & \textit{Low}  & \textit{High} & \textit{Overall} \\
\hline
Base network &36.6 & 94.7 & 60.1 &17.8& 103.5 &90.8& 39.2 & 95.9 & 61.9 &18.7& 106.5 &93.6  \\
w/ Cutout \citep{devries2017cutout} &35.9 & 93.6 & 58.9 &\textbf{15.7}&101.9 &88.6 & 38.6 & 95.0 & 61.2 &\textbf{16.5}& 104.7 &91.8  \\
w/ CutMix \citep{yun2019cutmix} &35.7 & 93.2 & 58.6 &16.4&99.6&87.9 &38.0 & 94.3 &60.6 &17.8&103.1 &90.6  \\
\rowcolor{Gray}
w/ ours  &\textbf{34.9} & \textbf{91.7} & \textbf{57.2} &16.8& \textbf{98.3} &\textbf{87.1} &\textbf{37.8} & \textbf{92.2} &\textbf{59.1} &18.3& \textbf{101.3} &\textbf{88.9} \\
\bottomrule
\end{tabular}
\end{threeparttable}}
\caption{\textbf{Effect of occlusion simulation.} Compared to a base network and our reimplementation of two related works, our occlusion augmentation reduces the count error (MAE) for the entire test set, the \emph{low} occlusion set, and especially for the \emph{high} occlusion set. 
}
\label{tab:occlusion}
\end{table*}
\begin{figure*}[!t]
\centering
\begin{subfigure}{0.244\textwidth}
\includegraphics[width=\textwidth]{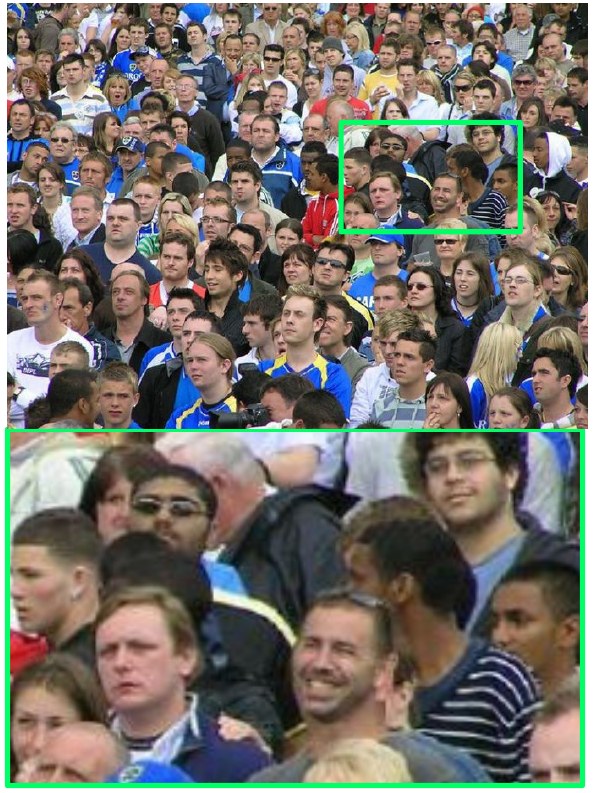}
\caption{\textbf{Original image}}
\label{fig:augmentation-a}    
\end{subfigure}
\begin{subfigure}{0.244\textwidth}
\includegraphics[width=\textwidth]{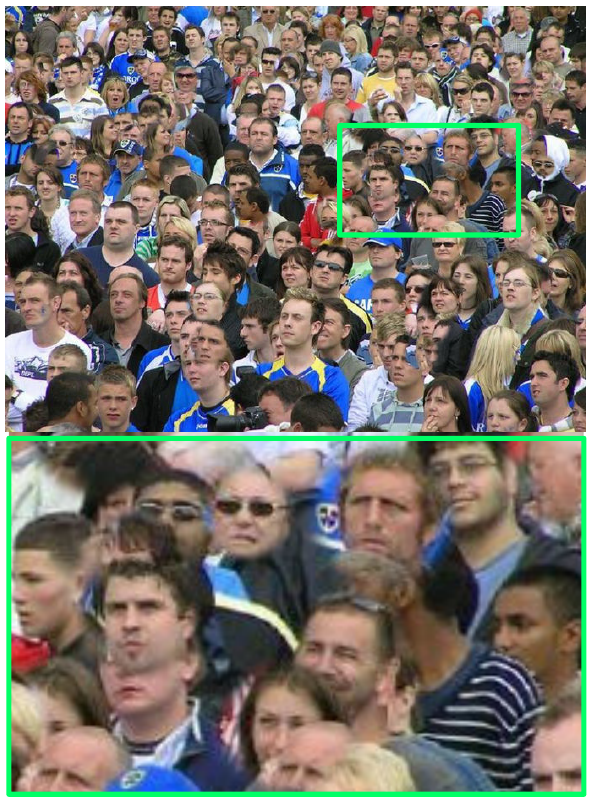}
\caption{\textbf{Our image}}
\label{fig:augmentation-b}    
\end{subfigure}
\begin{subfigure}{0.244\textwidth}
\includegraphics[width=\textwidth]{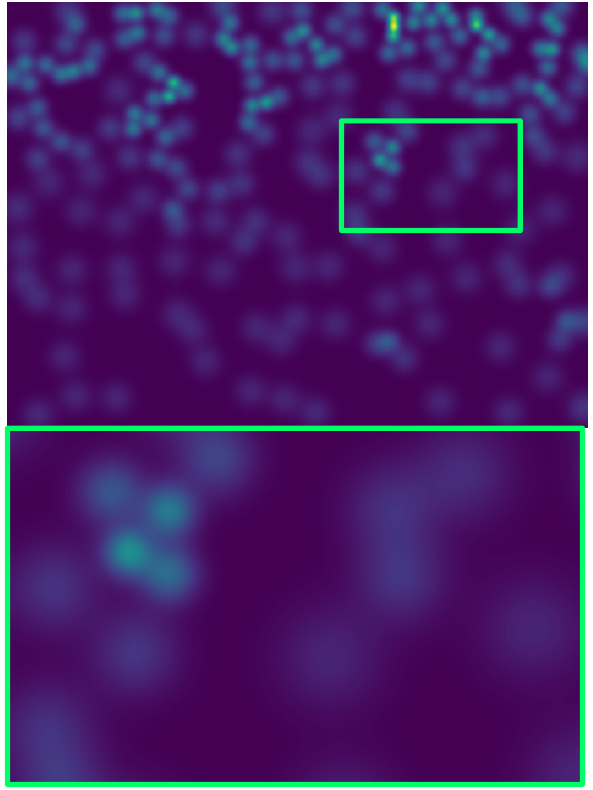}
\caption{\textbf{Original density}}
\label{fig:augmentation-c}    
\end{subfigure}
\begin{subfigure}{0.244\textwidth}
\includegraphics[width=\textwidth]{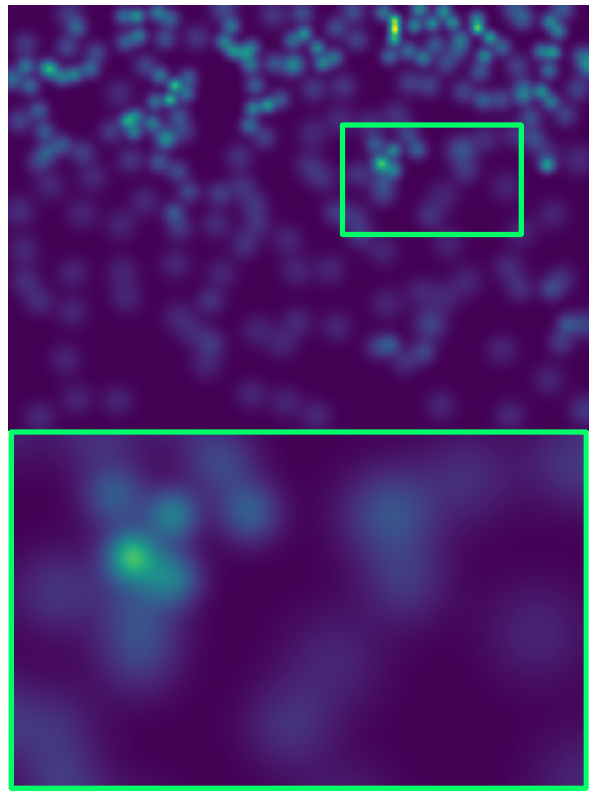}
\caption{\textbf{Our density}}
\label{fig:augmentation-d}    
\end{subfigure}
\caption{\textbf{Occlusion creation}. Illustrative examples for creating new occlusions on images and corresponding density maps. Our occlusion-augmented image and density look quite natural, as masked in the green regions.}
\label{fig:augmentation}   
\end{figure*}
%
%===============================
\subsection{Experimental setup}
%===============================
\textbf{Datsets.}  We consider five counting datasets in this paper, as commonly used in the recent literature, \eg,~\citep{bai2020adaptive,wan2021generalized,liu2021exploiting,wang2021uniformity,song2021rethinking,ma2021towards}:
\textbf{\textit{ShanghaiTech}} \citep{zhang2016single} consists of 1,198 images with 330,165 pedestrians. This dataset is divided into two parts: \textbf{\textit{Part\_A}} with 482 images in which crowds are mostly dense, and \textbf{\textit{Part\_B}} with 716 images, where crowds are sparser. Each part is divided into a training and testing subset as specified in \citep{zhang2016single}. 
\textbf{\textit{UCF-QNRF}} \citep{idrees2018composition} consists of 1,535 images, with the count ranging from 49 to 12,865. 
For training 1,201 images are available, and the remaining 334 form the test set. 
\textbf{\textit{JHU-CROWD++}} \citep{sindagi2020jhu-crowd++} consists of 4,372 images with a total of 1.51 million point annotations. 
The dataset is split into a training set of 2,272 images, a validation set of 500 images and a testing set of 1,600 images.
\textbf{\textit{NWPU-Crowd}} \citep{wang2020nwpu} consists of 5,109 images with over 2 million point annotations. 
The dataset is split into a training set of 3,109 images, a validation set of 500 images and a testing set of 1,500 images. 

For all datasets, we augment the images by randomly performing horizontal flipping, and randomly cropping $256 {\times} 256$ patches for ShanghaiTech Part\_A, $384 {\times} 384$ patches for JHU-CROWD++ and NWPU-Crowd, and $512 {\times} 512$ patches for ShanghaiTech Part\_B and UCF-QNRF. For Gaussian density map generation, we follow the previously suggested dataset settings from \citep{zhang2016single,idrees2018composition,sindagi2020jhu-crowd++}.

\textbf{Implementation details.}
We adopted T2T-Vit-14 \citep{yuan2021tokens} as our transformer backbone network and integrated our methods using the guidelines outlined in Section \ref{sec:network}. Additionally, we incorporated CSRNet \citep{Li_2018_CVPR} as our convolutional backbone network. CSRNet has been widely used in the literature, as demonstrated by its inclusion in multiple studies \citep{zhao2019leveraging,ma2019bayesian,wan2020modeling,wang2020distribution,wan2021generalized,ma2021towards}. To enhance its performance, we introduced batch normalization layers into our implementation of CSRNet, similar to the approach taken by \cite{bai2020adaptive}. This improved version is called CSRNet+. For our integration with CSRNet+, we used the same backbone network as the frontend network and the same segmentation focus and density regressor as the backend network. The global density  focus is the same as that used with T2T-Vit-14, but we employed a bilinear pooling layer instead of using the global density class token. Both networks have been initialized with their ImageNet pre-trained models and trained using Adam with batches of 10. By default, we have used a $\ell_1$ loss function. The learning rate has been fixed at 1e-4 for CSRNet+ and le-5 for T2T-Vit-14.

\textbf{Metrics.} We report the standardized Mean Absolute Error (MAE) and Root Mean Square Error (RMSE) metrics given count estimates and their ground-truth.
\begin{table*}[!t]
\centering
\resizebox{0.99\linewidth}{!}{
\begin{threeparttable}
\begin{tabular}{@{}lccccccccccccccc@{}}
\toprule
& \multicolumn{6}{c}{\textbf{\textbf{ T2T-Vit}}} & \multicolumn{6}{c}{\textbf{\textbf{CSRNet+}}}\\
\cmidrule(lr){2-7} \cmidrule(lr){8-13}
& \multicolumn{3}{c}{\textbf{Part\_A}} &
\multicolumn{3}{c}{\textbf{UCF\_QNRF}} & \multicolumn{3}{c}{\textbf{Part\_A}} &
\multicolumn{3}{c}{\textbf{UCF\_QNRF}} \\
\cmidrule(lr){2-4} \cmidrule(lr){5-7} \cmidrule(lr){8-10} \cmidrule(lr){11-13} 
& \textit{BG}  & \textit{FG}  & \textit{Overall} & \textit{BG} & \textit{FG} & \textit{Overall} & \textit{BG}  & \textit{FG}  & \textit{Overall} & \textit{BG} & \textit{FG} & \textit{Overall} \\
\hline
Base network & 3.2 & 59.1 & 60.1 &8.2 & 89.7 &90.8 & 3.4 & 61.5 & 61.9 &8.7 & 91.3 &93.6 \\
w/ standard distillation\citep{hinton2015distilling} & 3.1 & 58.8 &59.8 &8.1 & 89.1 &90.2 & 3.3 & 61.1 & 61.4 &8.6 & 91.0 &93.1  \\
w/ attention-injective \citep{liu2019adcrowdnet} & 3.0 & 58.2 &59.1 &5.8 & 88.5 &89.4 & 3.3 & 59.7 & 60.5 &6.1 & 89.9 &91.2 \\
w/ segmentation attention \citep{modolo2021understanding} & \textbf{2.8} & 57.7 &58.2 &4.4 & 87.6 &88.8 & \textbf{2.9} & 58.8 & 59.3 &4.5 & 88.3 &89.4  \\
w/ background feature filtering \citep{mo2020background} &2.9 & 57.1 &57.7 &4.8 & 86.7 &87.9 & 3.1 & 58.2 & 58.8 &4.3 & 87.1 &88.3  \\
w/ decoupled learning \citep{cheng2021decoupled} &3.0 & 56.3 &57.1 &\textbf{4.2} & \textbf{85.8} &\textbf{86.4} & 3.1 & 57.8 & 58.2 &3.5 &\textbf{86.2} &\textbf{87.3} \\
%\hline
\rowcolor{Gray}
w/ our foreground distillation &2.9& \textbf{55.7} &\textbf{56.5} &4.8 &86.3&86.7 & 3.0 & \textbf{57.1}&\textbf{57.6}&\textbf{3.4}& 86.8 &87.8 \\
\bottomrule
\end{tabular}
\end{threeparttable}}
\caption{\textbf{Effect of foreground distillation.} Compared to base networks and our reimplementation of five related works, our simple foreground distillation better reduces the overall count error (MAE), for both background (BG) and foreground (FG). 
}
\label{tab:seg+dis}
\end{table*}
\begin{figure*}[t!]
\centering
\begin{subfigure}{0.24\textwidth}
\includegraphics[width=\textwidth]{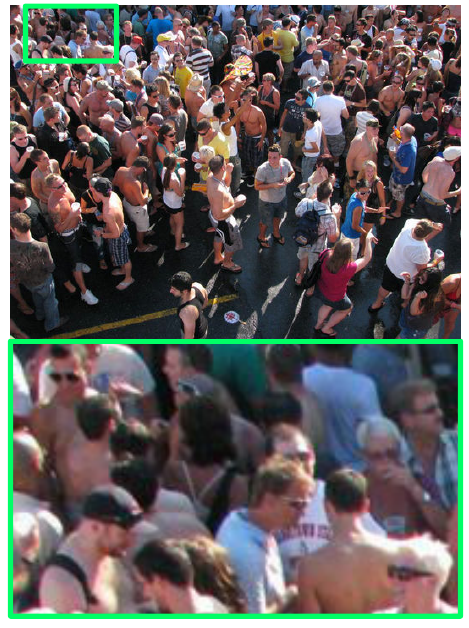}
\caption{{\scriptsize \textbf{Training image}}}
\label{fig:augmentation-a}    
\end{subfigure}
\begin{subfigure}{0.24\textwidth}
\includegraphics[width=\textwidth]{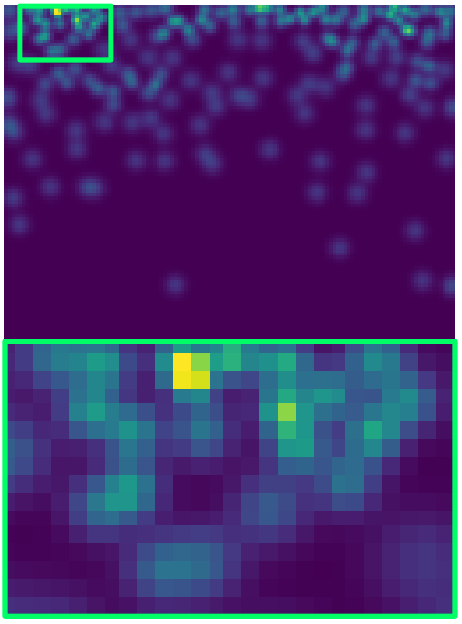}
\caption{{\scriptsize \textbf{Gaussian density}}}
\label{fig:augmentation-b}    
\end{subfigure}
\begin{subfigure}{0.245\textwidth}
\includegraphics[width=\textwidth]{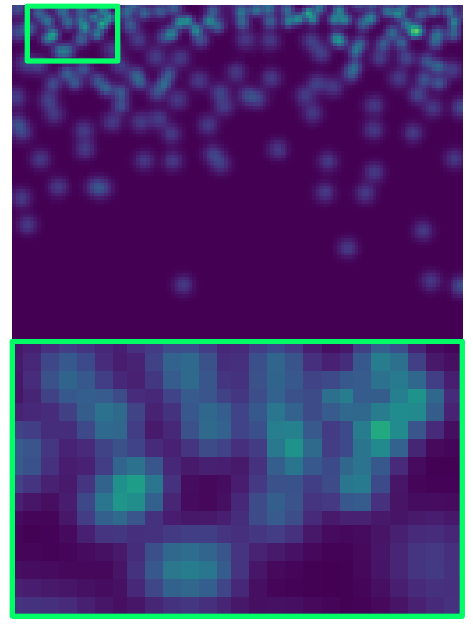}
\caption{{\scriptsize \textbf{Distilled density}}}
\label{fig:augmentation-c}    
\end{subfigure}
\begin{subfigure}{0.242\textwidth}
\includegraphics[width=\textwidth]{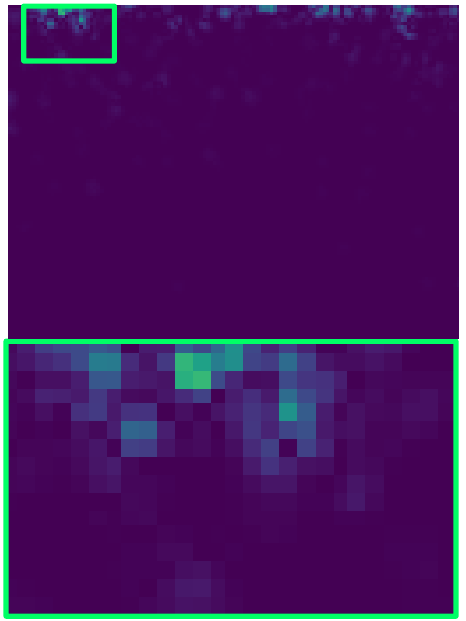}
\caption{{\scriptsize \textbf{Difference}}}
\label{fig:augmentation-d}    
\end{subfigure}
\caption{\textbf{Density maps from distillation}. Illustrative examples of Gaussian and distilled density maps. Compared to Gaussian densities, distilled densities better reflect the true foreground knowledge, reducing local noisy maxima (yellow region) and matching the objects in the image, especially occluded objects.}
\label{fig:distillation}    
\end{figure*}
%
%===============================
\subsection{Counting-specific occlusion simulation}
%===============================
We first demonstrate the potential of our occlusion handling augmentation. To highlight its effectiveness in highly occluded images, we divide the test images into two sets according to their occlusion level, as computed by Equation \ref{eq:occmap}. The test images are grouped into the \emph{low} occlusion set if their occlusion level is lower than $1.5$, the remaining images are grouped into the \emph{high} occlusion set. For ShanghaiTech Part\_A, we obtain 109 low and 73 high occlusion images and for UCF\_QNRF, we obtain 49 low and 285 high occlusion images.

To highlight that the key here is occlusions, not just the augmentation, we have additionally investigated the effect of adding well-known occlusion augmentation approaches such as Cutout \citep{devries2017cutout} and CutMix \citep{yun2019cutmix}. Cutout randomly zero-outs a region in the training images. Instead of simply removing pixels, CutMix replaces the removed regions with a patch from another image. Cutout and CutMix create new training samples, but they do not create new occluded samples. Differently, our approach creates a variety of occluded samples by explicitly simulating the occlusion in the real world.

The results are shown in Table \ref{tab:occlusion}. Compared to the T2T-Vit base counting network, both Cutout and CutMix reduce the counting error, but cannot compete with our occlusion simulation (Part\_A: 58.9~for Cutout, 58.6 for CutMix, and 57.2 for ours - UCF\_QNRF: 88.6~for Cutout, 87.9 for CutMix, and 87.1 for ours). Especially, our approach works much better when occlusion levels are high (Part\_A: 93.6~for Cutout, 93.2~for CutMix, and 91.7 for ours - UCF\_QNRF: 101.9~for Cutout, 99.6~for CutMix, and 98.3 for ours). This experiment solidifies our point: occlusions weight heavily on the counting error, and there are not enough natural occlusions to learn from. Instead, by simulating occlusions, we can easily reduce the MAE effectively.
%
%===============================
\subsection{Foreground distillation}
\label{sec:exp_segmentation}
%===============================
In this experiment, we evaluate the effectiveness of foreground distillation. We compare it to the standard distillation \citep{hinton2015distilling} on top of the same counting network, where we use the same network architecture for both student and teacher for a fair comparison. We also compare it to the other four baselines that were proposed for reducing background errors as well. The first baseline \citep{liu2019adcrowdnet} is a cascade counting approach, which cascades a crowd-level attention prediction network and a density estimation network. The second baseline \citep{modolo2021understanding} reduces the background error by learning a segmentation attention map for the output density map. The third one \citep{mo2020background} filters the background noise on the intermediate features by learning a head segmentation mask. The fourth baseline \citep{cheng2021decoupled} learns to count with decoupled learning, which first regresses a probability map to identify possible object regions, and then predicts a density map on the probability map to count the objects. All baselines use the same backbone and training optimizations as our method for a fair comparison. 

As shown in Table~\ref{tab:seg+dis}, the standard distillation just marginally improves the counting performance. For instance, on ShanghaiTech Part\_A, the MAE reduces from 60.1 to 59.8, and on UCF\_QNRF, it decreases from 90.8 to 90.1, when using the T2T-Vit base network. However, all four other baselines outperform standard distillation. Notably, our proposed method surpasses the performance of these four baselines (59.1~\citep{liu2019adcrowdnet}, 58.2~\citep{modolo2021understanding}, 57.7~\citep{mo2020background}, 57.1~\citep{cheng2021decoupled}) on ShanghaiTech Part\_A, achieving an impressive MAE of 56.5. On UCF\_QNRF, our method outperforms all baselines except the fourth one \citep{cheng2021decoupled}. It is worth mentioning that \cite{cheng2021decoupled} utilizes a two-network cascade for improved results, albeit at the cost of inefficient inference. Furthermore, we also observe a substantial reduction in MAE when applying our proposed method with the CSRNet+ base network. In summary, foreground distillation emerges as a straightforward yet effective approach, adaptable to any network, and capable of substantially enhancing counting accuracy.

In Figure \ref{fig:distillation}, we provide some examples of Gaussian and distilled density maps. We observe that distilled densities better reflect foreground counting knowledge, matching the observed objects, especially the occluded objects in the image, compared to the Gaussian densities. Thus, distillation provides a more natural relation between the foreground of the images and density maps.
\begin{table}[!t]
\centering
\resizebox{\columnwidth}{!}{
\begin{threeparttable}
\begin{tabular}{@{}lccccccccc@{}}
\toprule
& \multicolumn{2}{c}{\textbf{\textbf{T2T-Vit}}} & \multicolumn{2}{c}{\textbf{\textbf{CSRNet+}}}\\
\cmidrule(lr){2-3} \cmidrule(lr){4-5}
& \textbf{Part\_A} &
\textbf{QNRF} & \textbf{Part\_A} &
\textbf{QNRF} \\
\hline
Base network &60.1&90.8 &61.9 &93.6\\
w/ \cite{chen2017sca} &59.2&89.7 &61.2 &91.8\\
\rowcolor{Gray}
w/ local segmentation focus &58.4&88.3 &59.4 &90.6\\
\hline
w/ SE \cite{woo2018cbam} &59.7&89.3 &61.2 &91.8\\
w/CBAM \cite{hu2018squeeze} &59.4&88.9 & 60.6 &90.5 \\
\rowcolor{Gray}
w/ global density focus &58.7&87.6 &59.1 &88.6\\
\hline
\rowcolor{Gray}
w/ combined focus for free &56.9&85.8 &57.6 &86.4\\
\bottomrule
\end{tabular}
\end{threeparttable}}
\caption{\textbf{Effect of local and global focus for free.} Our focus for free consisits of  focus from segmentation and focus from global density, which incorporates explicit supervisions for spatial attention and channel attention, outperforms both the base networks and three related works that implement spatial attention and channel attention implicitly. When combined, our focus for free further reduces MAE significantly across datasets and architectures. 
}
\label{tab:focus}
\end{table}
\begin{table*}[!t]
\centering
\resizebox{0.99\linewidth}{!}{
\begin{tabular}{ccccccccc}
\toprule
\multicolumn{3}{c}{\textbf{Three methods of repurposing point annotations}} & \multicolumn{2}{c}{\textbf{T2T-Vit}}&\multicolumn{2}{c}{\textbf{CSRNet+}} \\
\cmidrule(lr){1-3} \cmidrule(lr){4-5} \cmidrule(lr){6-7}
Occlusion simulation & Foreground distillation & Focus for free &\textbf{Part\_A} & \textbf{UCF\_QNRF} &\textbf{Part\_A} & \textbf{UCF\_QNRF}  \\
\hline
 & & &60.1 &90.8& 61.9 & 93.6 \\
 \checkmark & & &57.2 &87.1& 59.1&88.9\\
 \checkmark&\checkmark &  &55.8 &85.5&56.2 &86.2\\
 \checkmark& & \checkmark &56.3 &84.9&56.8 &85.6\\
&\checkmark & \checkmark &52.2 &82.6& 53.5 &83.7\\
\rowcolor{Gray}
\checkmark&\checkmark & \checkmark &\textbf{51.6} & \textbf{80.9} & \textbf{52.7} & \textbf{81.8} \\
\bottomrule
\end{tabular}
}
\caption{\textbf{Effect of combining our three methods} in terms of MAE on ShanghaiTech Part\_A and UCF\_QNRF. With each new addition, the counting error decreases, showing their complementary nature.}
\label{tab:combine}
\end{table*}
\begin{table*}[t!]
\centering
\resizebox{0.99 \linewidth}{!}{
\begin{tabular}{@{}lccccccccccccr@{}}
\toprule
&\multirow{2}{*}{\textbf{Backbone}} &\multicolumn{2}{c}{\textbf{Part\_A}} & \multicolumn{2}{c}{\textbf{Part\_B}}&\multicolumn{2}{c}{\textbf{UCF\_QNRF}}&\multicolumn{2}{c}{\textbf{JHU-CROWD++}}&\multicolumn{2}{c}{\textbf{NWPU-Crowd}}\\
\cmidrule(lr){3-4} \cmidrule(lr){5-6} \cmidrule(lr){7-8} \cmidrule(lr){9-10}  \cmidrule(lr){11-12}
 & & MAE & RMSE & MAE & RMSE & MAE & RMSE & MAE & RMSE & MAE &  RMSE\\
\hline
\rowcolor{Gray} 
\textbf{$\ell_1$/$\ell_2$-norm loss}&&&&&&&&&&&&\\
\cite{xu2022autoscale} &VGG-16&65.8 &112.1 &8.6 &13.9 &104.4 &174.2 &76.4&292.7&94.1&388.2\\
\cite{wan2020kernel} &VGG-16&63.8 &99.2 &7.8 &12.7 &99.5 &173.0 &69.7&268.3&100.5&415.5\\
\cite{ma2021towards} &VGG-16&58.4 &97.9 &- &- &96.3&155.7 &65.1&269.3 &- &-\\
\cite{shu2022crowd} &VGG19&57.5&94.3 &6.9 &11.0 &80.3 &137.6 &57.0 &235.7 &\textbf{76.8} &\textbf{343.0} \\
\cite{cheng2021decoupled} &VGG-16&57.2 &93.0 &\underline{6.3} &10.7 &81.7&137.9 &73.7&292.5 &85.5 &361.5\\
\cite{bai2020adaptive} &VGG-16&55.4 &97.7 &6.4 &11.3 &\textbf{71.3}&132.5&- &- &- &-\\
\cite{tran2022improving} &ViT-B/32&54.8 &\textbf{80.9} &8.6 &13.8 &87.0&141.9 &-&- &- &-\\
\cite{cheng2022rethinking} &ResNet-50&54.8 &89.1 &\textbf{6.2} &\textbf{9.9} &81.6&153.7 &58.2&\textbf{245.1} &- &-\\
\cite{wang2021uniformity} &VGG-16&54.6 &91.2 &6.4 &10.9 &81.1&\underline{131.7} &-&- &- &-\\
\textit{\textbf{Ours (CSRNet+)}} &VGG-16&\underline{52.7} &88.6 &6.9 &11.2 &81.8&133.5 &\underline{57.9}&258.4 &78.5 &347.8\\
\textit{\textbf{Ours (T2T-Vit)}} &T2T-Vit-14&\textbf{51.6}&\underline{84.2} &6.5 &\underline{10.3} &\underline{80.9}& \textbf{129.5} &\textbf{56.8}&\underline{254.6} &\underline{77.4}&\underline{345.9}\\
\midrule
\rowcolor{Gray} 
\textbf{DM loss}&&&&&&&&&&&&\\
\cite{wan2021generalized} &VGG19&61.3 &95.4 &7.3 &11.7 &84.3&147.5 &59.9&259.5 &79.3 &346.1\\
\cite{wang2020distribution}  &VGG-19&59.7 &95.7 &7.4 &11.8 &85.6 &148.3 &- &- &88.4 &388.6 \\
\cite{liu2021exploiting} &ResNet-18&55.4 &91.3 &6.9 &10.3 &\textbf{76.2}&\textbf{121.5} &59.9&259.5 &\textbf{74.7} &\textbf{267.9}\\
\cite{ma2021towards} &VGG-19&55.0 &92.7 &- &- &80.7&146.3&59.3&248.9 &- &-\\
\cite{sun2021boosting} &T2T-Vit-14&53.1 &\textbf{82.2} &7.3 &11.5 &83.8&143.4 &54.8&\textbf{208.5} &82.0 &366.9\\
\cite{song2021rethinking} &VGG-16&52.7 &85.1 &6.3 &\textbf{9.9} &85.3&154.5 &-&- &77.4 &362.0\\
\textit{\textbf{Ours (CSRNet+)}} &VGG16&\underline{51.2} &85.7 &6.7 &10.9 &80.2&131.7 &56.4&252.4 &77.9 &349.8\\
\textit{\textbf{Ours (T2T-Vit)}} &T2T-Vit-14&\textbf{50.4}&\underline{84.2}&\textbf{6.2} &\underline{10.1} &\underline{78.4}& \underline{130.9} &\textbf{54.1}&\underline{248.3}&\underline{76.1} &\underline{330.5}\\

\bottomrule
\end{tabular}}
\caption{\textbf{Comparison to the state-of-the-art} on ShanghaiTech Part\_A, Part\_B, UCF\_QNRF, JHU-CROWD++ and NWPU-Crowd.
%Our results set a new state-of-the-art on most datasets for most metrics.
We combine our three methods with CSRNet+ and T2T-Vit to obtain counting results that are either competitive or better than the current state-of-the-art, highlighting their potential for counting. \textbf{Bold} indicates the best counting results, and \underline{underline} indicates the second-best counting results.}
\label{tab:sota}
\end{table*}

%===============================
\subsection{Local and global focus}
%===============================
Next, we demonstrate the effect of our proposed focus for free that includes focus from segmentation and focus from global density. We first compare each to their respective baseline approaches. Then, we report the results of the combination of the two methods.

\textbf{Local segmentation focus.} We first analyze the effect of the proposed segmentation focus. We compare to two baselines. The first performs counting using the base network, where the loss is only optimized with respect to the density map estimation. The second baseline adds spatial attention on top of this base network, as proposed in \cite{chen2017sca}. The results are shown in Table \ref{tab:focus}. For ShanghaiTech Part\_A, the T2T-Vit base network obtains an MAE of 60.1 and the addition of spatial-attention actually increases the count error to 59.2 MAE, as it fails to emphasize relevant features. In contrast, our proposed focus from segmentation can explicitly guide the network to focus on task-relevant regions, and it reduces the count error from 60.1 to 58.4 MAE.

\textbf{Global density focus.} Next, we demonstrate the effect of our proposed global density focus. For this experiment, we again compare it with the SE module \citep{hu2018squeeze} and the channel attention module of CBAM \citep{woo2018cbam}, while keeping the base network unchanged. To ensure a fair comparison, we use the same MLP network for all the approaches. The results of our experiment are presented in Table \ref{tab:focus}. Both SE and CBAM can already improve the base network's performance based on both CSRNet+ and T2T-Vit. Our proposed method outperforms SE and CBAM, reducing the count error even further, due to its explicit supervision. 

\textbf{Combined focus for free.}
In the aforementioned experiments, we have shown that each focus matters for counting. In this experiment, we combine these two focuses for more accurate counting, in view that these two focuses aid density map estimation respectively from a local and global perspective, complementing each other. The results are shown in Table \ref{tab:focus}. The combination achieves a reduced count error of 56.9 MAE on ShanghaiTech Part\_A, and obtains a reduced count error of 85.8 MAE on UCF\_QNRF.

%===============================
\subsection{Comparative evaluation} \label{exp}
%===============================
For the final experiments, we first show the effect of combining the three methods we propose for improving counting by repurposing point annotations. Then we compare to the state-of-the-art in counting.

\textbf{Combining the three approaches.} Until now, we have established that every method plays a role in counting on its own. In this study, we aim to assess whether these methods are also complementary to each other. The results of this experiment are shown in Table \ref{tab:combine}, utilizing both T2T-Vit and CSRNet+ as backbones. Compared to standard counting with just the base networks, we reduce the MAE considerably by handling occluded objects with occlusion simulation (Section \ref{sec:occlusion}). By combining occlusion simulation with foreground distillation (Section \ref{sec:segmentation}) or focus for free (Section \ref{sec:focus}), or combining foreground distillation with focus for free, the MAE is further reduced. Combining all three solutions achieves the lowest count error across datasets and base networks.
\begin{figure*}[h!]
\centering
\begin{subfigure}{0.244\textwidth}
\includegraphics[width=\textwidth]{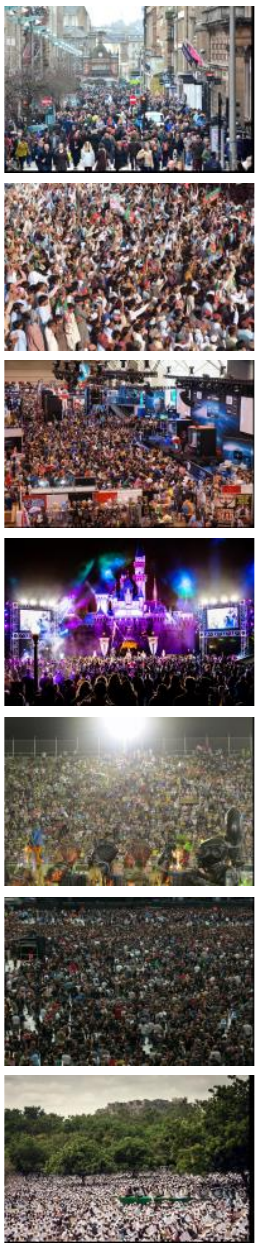}
\caption{\textbf{Input image}}
\label{fig:case-a}    
\end{subfigure}
\begin{subfigure}{0.241\textwidth}
\includegraphics[width=\textwidth]{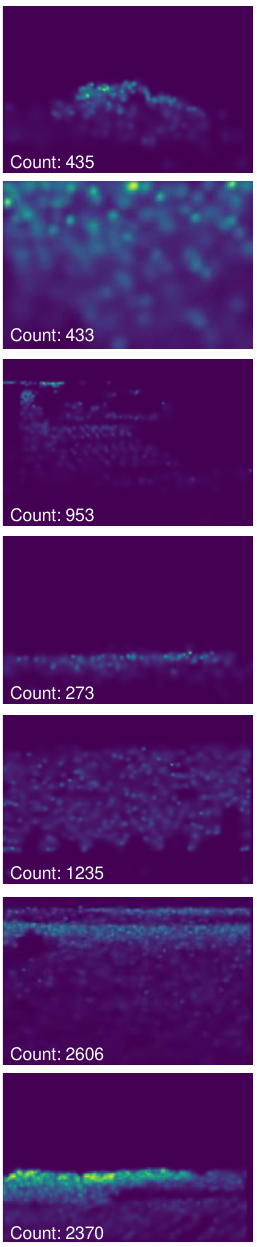}
\caption{\textbf{Ground-truth}}
\label{fig:case-b}    
\end{subfigure}
\begin{subfigure}{0.244\textwidth}
\includegraphics[width=\textwidth]{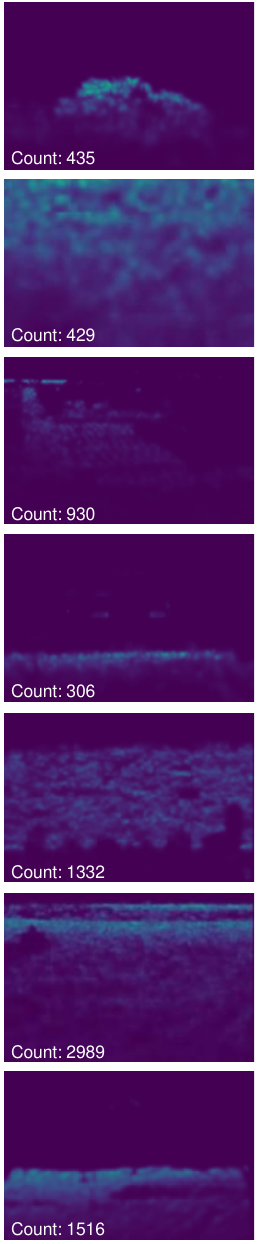}
\caption{\textbf{Ours (T2T-Vit)}}
\label{fig:case-c}    
\end{subfigure}
\begin{subfigure}{0.243\textwidth}
\includegraphics[width=\textwidth]{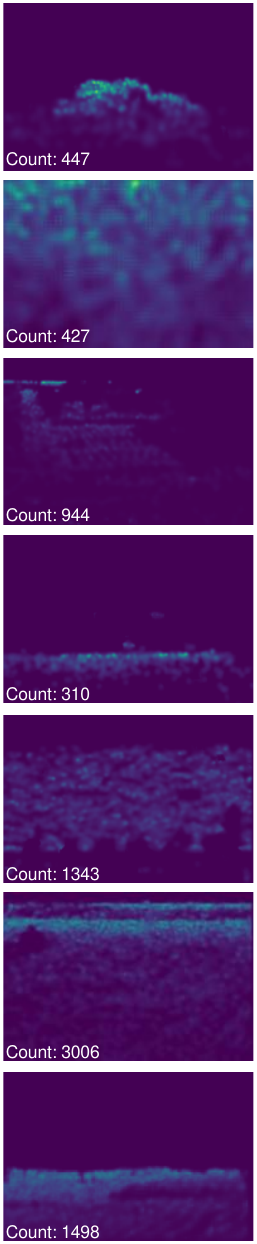}
\caption{\textbf{Ours (CSRNet+)}}
\label{fig:case-d}    
\end{subfigure}
\caption{\textbf{Success and failure cases obtained with $\ell_1$-norm loss.} When objects are individually visible, we can count them accurately (first four rows). Further improvements are required for extremely dense scenes where individual objects are hard to distinguish, or where objects blend with the context (last three rows). }
\label{fig:cases-l1}   
\end{figure*}
\begin{figure*}[h!]
\centering
\begin{subfigure}{0.242\textwidth}
\includegraphics[width=\textwidth]{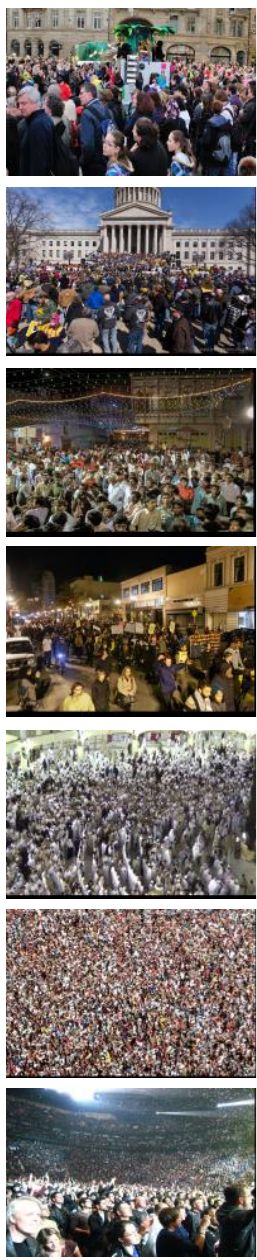}
\caption{\textbf{Input image}}
\label{fig:case-a}    
\end{subfigure}
\begin{subfigure}{0.242\textwidth}
\includegraphics[width=\textwidth]{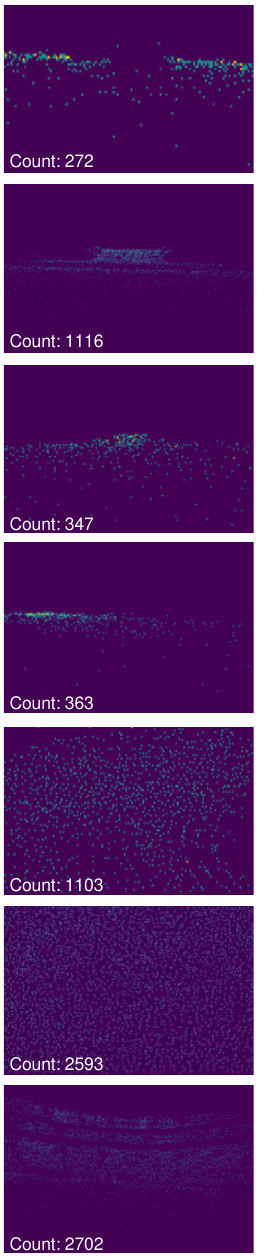}
\caption{\textbf{Ground-truth}}
\label{fig:case-b}    
\end{subfigure}
\begin{subfigure}{0.242\textwidth}
\includegraphics[width=\textwidth]{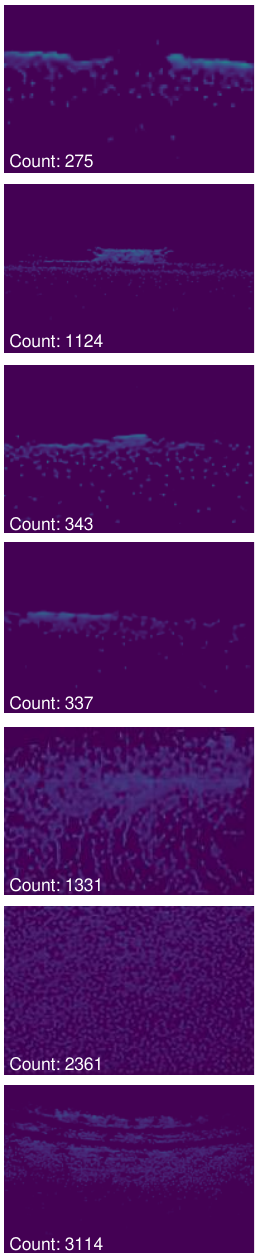}
\caption{\textbf{Ours (T2T-Vit)}}
\label{fig:case-c}    
\end{subfigure}
\begin{subfigure}{0.239\textwidth}
\includegraphics[width=\textwidth]{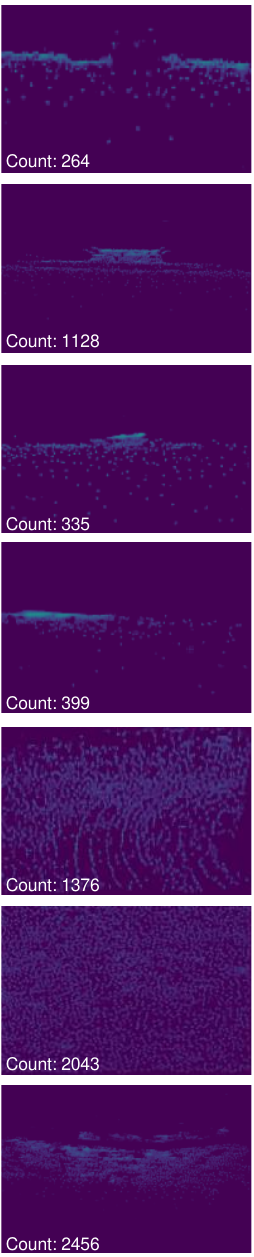}
\caption{\textbf{Ours (CSRNet+)}}
\label{fig:case-d}    
\end{subfigure}
\caption{\textbf{Success and failure cases obtained with DM loss.} When objects are individually visible, we can count them accurately (first four rows). Further improvements are required for extremely dense scenes where individual objects are hard to distinguish, or where objects blend with the context (last three rows). }
\label{fig:cases-dm}    
\end{figure*}

\textbf{Comparison to the state-of-the-art.}
As the ultimate test, we draw a comparison to the state-of-the-art in counting on five datasets.
We report our results using both CSRNet+ and T2T-Vit as base networks, and both $\ell_1$ and distribution matching (DM) \citep{wang2020distribution} as loss functions. Table \ref{tab:sota} shows the comparative evaluation over all datasets and metrics. The results show that despite using canonical counting backbones, which on their own can not compete with the state-of-the-art, we obtain the best or second-best counting results on 8 of the 10 metrics when using $\ell_1$ loss. This highlights the potential of our simple solutions. We also compare to recent counting alternatives that utilize the advanced DM loss and find that we maintain the best or second-best results for all metrics. We conclude that with our proposed methods, we can make further reduce the counting errors.
We show some success and failure results obtained by our methods in Fig.~\ref{fig:cases-l1} and \ref{fig:cases-dm}. Even in challenging scenes with relatively sparse small objects or relatively dense large objects, our method is able to achieve an accurate count (first four rows). Our approach fails when dealing with extremely dense scenes where individual objects are hard to distinguish, or where objects blend with the context (last three rows). Such scenarios remain open counting challenges.

\section{Conclusion}
This paper aims to enhance density-based counting by leveraging point annotations beyond their original purpose of creating a density map. We propose occlusion augmentation, foreground distillation, and focus for free. Foreground distillation involves generating foreground masks from point annotations and utilizing these masks to distill foreground knowledge. By doing so, this approach mitigates the influence of background pixels on counting accuracy. Counting-specific occlusion augmentation leverages point annotations to simulate occluded objects in both the input and density images, enabling the network to become more robust to occlusions. This approach improves the network's ability to handle challenging and crowded scenarios where objects may be partially or fully overlapping. Additionally, we utilize explicit supervision derived from point annotations to offer both local and global focus. This approach ensures that the network pays attention to crucial features in the counting process. These approaches work complementary to improve counting accuracy and can be seamlessly integrated into various network architectures while accommodating different loss functions. The effectiveness of the methods is evaluated through experiments on five benchmark datasets. The results demonstrate that the proposed approaches consistently achieve remarkable counting results across various scenes. However, the work also highlights that counting in extremely dense scenes still remains an open problem. This suggests that further advancements are necessary to attain perfect counting in such challenging scenarios.

%\begin{acknowledgements}
%If you'd like to thank anyone, place your comments here
%and remove the percent signs.
%\end{acknowledgements}

% Authors must disclose all relationships or interests that 
% could have direct or potential influence or impart bias on 
% the work: 
%
% \section*{Conflict of interest}
%
% The authors declare that they have no conflict of interest.

% BibTeX users please use one of
\bibliographystyle{spbasic}      % basic style, author-year citations
\bibliography{egbib}   % name your BibTeX data base

\end{document}